\documentclass[11pt]{article}
\usepackage{coling2018}
\usepackage{times}
\usepackage{url}
\usepackage{latexsym}
\usepackage{graphicx}
\usepackage{svg}
\usepackage{color}

\title{A Survey of Domain Adaptation for Neural Machine Translation}

\author{Chenhui Chu \\
Institute for Datability Science \\
  Osaka University \\
  {\tt chu@ids.osaka-u.ac.jp} \\\And
  Rui Wang \\
  National Institute of Information \\
  and Communications Technology \\
  {\tt wangrui@nict.go.jp} \\}

\date{}

\begin{document}
\maketitle
\begin{abstract}
Neural machine translation (NMT) is a deep learning based approach for machine translation, which yields the state-of-the-art translation performance in scenarios where large-scale parallel corpora are available. Although the high-quality and domain-specific translation is crucial in the real world, domain-specific corpora are usually scarce or nonexistent, and thus vanilla NMT performs poorly in such scenarios. Domain adaptation that leverages both out-of-domain parallel corpora as well as monolingual corpora for in-domain translation, is very important for domain-specific translation. In this paper, we give a comprehensive survey of the state-of-the-art domain adaptation techniques for NMT.
\end{abstract}

\section{Introduction}
\label{sec:intro}
%
\blfootnote{
 \hspace{-0.65cm}  
This work is licensed under a Creative Commons Attribution 4.0 International License. License details: \url{http://creativecommons.org/licenses/by/4.0/}
}


 Neural machine translation (NMT) 
\cite{DBLP:journals/corr/ChoMGBSB14,DBLP:journals/corr/SutskeverVL14,DBLP:journals/corr/BahdanauCB14} 
allows for end-to-end training of a translation system without the need to deal with word alignments, translation rules and complicated decoding algorithms, which are characteristics of statistical machine translation (SMT) systems \cite{koehn-EtAl:2007:PosterDemo}. 
NMT yields the state-of-the-art translation performance in resource rich scenarios \cite{bojar-EtAl:2017:WMT1,nakazawa-EtAl:2017:WAT2017}. 
However, currently, high quality parallel corpora of sufficient size are only available for a few language pairs such as languages paired with English and several European language pairs. Furthermore, for each language pair the sizes of the domain specific corpora and the number of domains available are limited. As such, for the majority of language pairs and domains, only few or no parallel corpora are available. It has been known that both vanilla SMT and NMT perform poorly for domain specific translation in low resource scenarios \cite{duh-EtAl:2013:Short,sennrich-schwenk-aransa:2013:ACL2013,DBLP:conf/emnlp/ZophYMK16,koehn-knowles:2017:NMT}. 

High quality domain specific machine translation (MT) systems are in high demand whereas general purpose MT has limited applications. 
In addition, general purpose translation systems usually perform poorly and hence it is important to develop translation systems for specific domains \cite{koehn-knowles:2017:NMT}. Leveraging out-of-domain parallel corpora and in-domain monolingual corpora to improve in-domain translation is known as domain adaptation for MT \cite{wang-EtAl:2016:COLING5,Chu:2018:JIP}.
For example, the Chinese-English patent domain parallel corpus has 1M sentence pairs \cite{conf-ntcir-GotoCLST13},  but for the spoken language domain parallel corpus there are only 200k sentences available \cite{cettolo2015iwslt}. 
MT typically performs poorly in a resource poor or domain mismatching scenario and thus it is important to leverage the spoken language domain data with the patent domain data \cite{chu:2017:ACL}. 
Furthermore, there are monolingual corpora containing millions of sentences for the spoken language domain, which can also be leveraged \cite{sennrich-haddow-birch:2016:P16-11}. 

There are many studies of domain adaptation for SMT, which can be mainly divided into two categories: data centric and model centric. Data centric methods focus on either selecting training data from out-of-domain parallel corpora based a language model (LM) \cite{moore-lewis:2010:Short,axelrod-he-gao:2011:EMNLP,duh-EtAl:2013:Short,hoang-simaan:2014:Coling,joty2015using,chen2016bilingual} or generating pseudo parallel data \cite{utiyama-isahara:2003:ACL,wang-EtAl:2014:EMNLP20142,chu:2015,wang-EtAl:2016:COLING5,marie-fujita:2017:Short}. Model centric methods interpolate in-domain and out-of-domain models in either a model level \cite{sennrich-schwenk-aransa:2013:ACL2013,joty2015using,Imamura:DomainAdaptation2016-1} or an instance level \cite{matsoukas-rosti-zhang:2009:EMNLP,foster-goutte-kuhn:2010:EMNLP,shah2010translation,8211,zhou2015domain}.
However, due to the different characteristics of SMT and NMT, many methods developed for SMT cannot be applied to NMT directly.

Domain adaptation for NMT is rather new and has attracted plenty of attention in the research community.
In the past two years, NMT has become the most popular MT approach and many domain adaptation techniques have been proposed and evaluated for NMT. These studies either borrow ideas from previous SMT studies and apply these ideas for NMT, or develop unique methods for NMT. Despite the rapid development in domain adaptation for NMT, there is no single compilation that summarizes and categorizes all approaches. As such a study will greatly benefit the community, we present in this paper a survey of all prominent domain adaptation techniques for NMT. There are survey papers for NMT \cite{DBLP:journals/corr/Neubig17,DBLP:journals/corr/abs-1709-07809}; however, they focus on general NMT and more diverse topics. Domain adaptation surveys have been done in the perspective of computer vision \cite{DBLP:journals/corr/Csurka17} and machine learning  \cite{Pan:2010:STL:1850483.1850545,Weiss2016}. However, such survey has not been done for NMT. To the best of our knowledge, this is the first comprehensive survey of domain adaptation for NMT.

\begin{figure}
\centering
\includegraphics[width=\hsize]{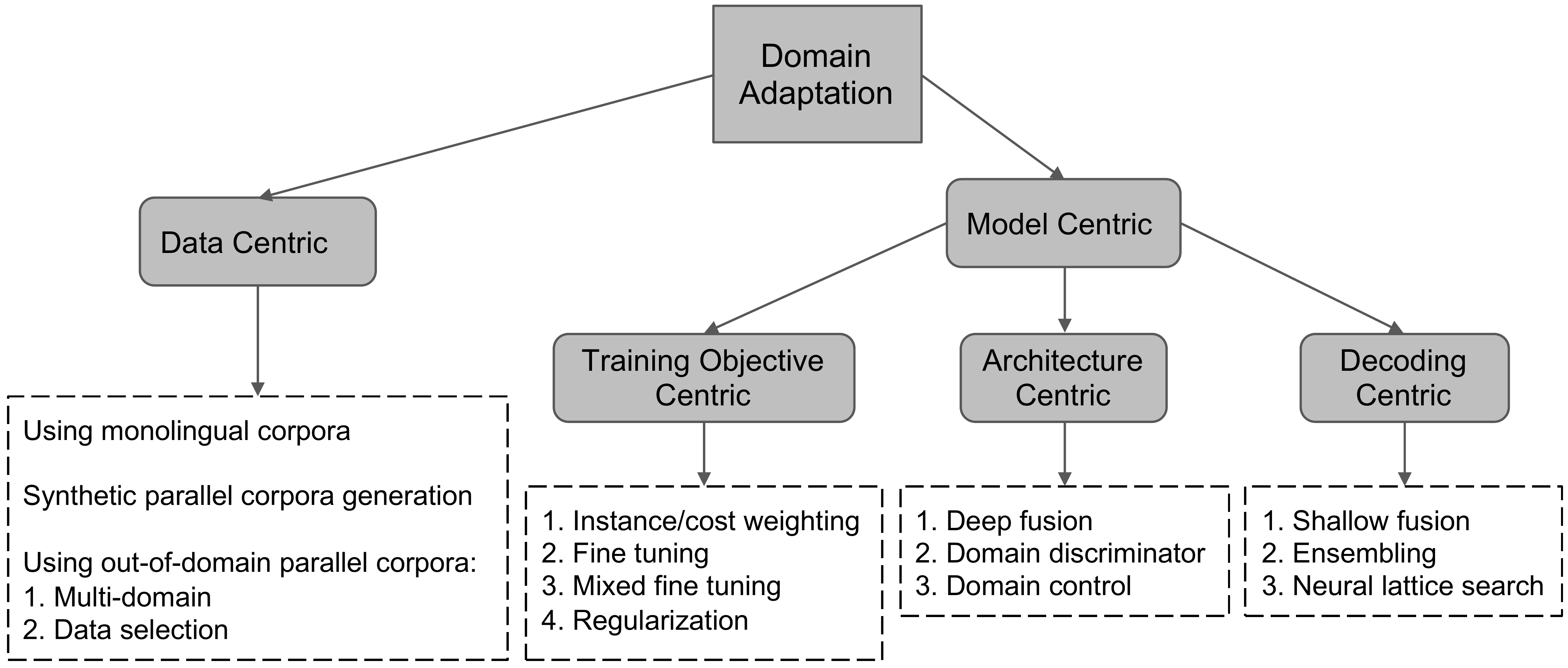}
\caption{Overview of domain adaptation for NMT.}
\label{fig:overview}
\end{figure}

In this paper, similar to SMT, we categorize domain adaptation for NMT into two main categories: data centric and model centric. The data centric category focuses on the data being used rather than  specialized models for domain adaptation. The data used can be either in-domain monolingual corpora \cite{zhang-zong:2016:EMNLP2016,cheng-EtAl:2016:P16-1,currey-micelibarone-heafield:2017:WMT,domhan-hieber:2017:EMNLP2017}, synthetic corpora \cite{sennrich-haddow-birch:2016:P16-11,zhang-zong:2016:EMNLP2016,DBLP:journals/corr/ParkSY17}, or parallel copora \cite{chu:2017:ACL,DBLP:journals/corr/abs-1708-08712,britz-le-pryzant:2017:WMT,wang-EtAl:2017:Short3,vanderwees-bisazza-monz:2017:EMNLP2017}. On the other hand, the model centric category focuses on NMT models that are specialized for domain adaptation, which can be either the training objective \cite{luong2015stanford,sennrich-haddow-birch:2016:P16-11,domspec,domfast,iwnmt,costweighting,varga:2017,dakw:17fine,chu:2017:ACL,micelibarone-EtAl:2017:EMNLP2017}, the NMT architecture \cite{domcont,DBLP:journals/corr/GulcehreFXCBLBS15,britz-le-pryzant:2017:WMT} or the decoding algorithm \cite{DBLP:journals/corr/GulcehreFXCBLBS15,dakw:17fine,khayrallah-EtAl:2017:I17-2}. An overview of these two categories is shown in Figure \ref{fig:overview}. Note that as model centric methods also use either monolingual or parallel corpora, there are overlaps between these two categories.

The remainder of this paper is structured as follows: We first give a brief introduction of NMT, and describe the reason for the difficulty of low resource domains and languages in NMT (Section \ref{sec:nmt}); Next, we briefly review the historical domain adaptation techniques being developed for SMT (Section \ref{sec:da-smt}); Under these background knowledge, we then present and compare the domain adaptation methods for NMT in detail (Section \ref{sec:da-nmt}); After that, we introduce domain adaptation for NMT in real word scenarios, which is crucial for the practical use of MT (Section \ref{sec:real}); Finally, we give our opinions of future research directions in this field (Section \ref{sec:future}) and conclude this paper (Section \ref{sec:conl}).

\section{Neural Machine Translation}
\label{sec:nmt}

 \begin{figure}[t]
     \begin{center}
          \includegraphics[width=0.8\hsize]{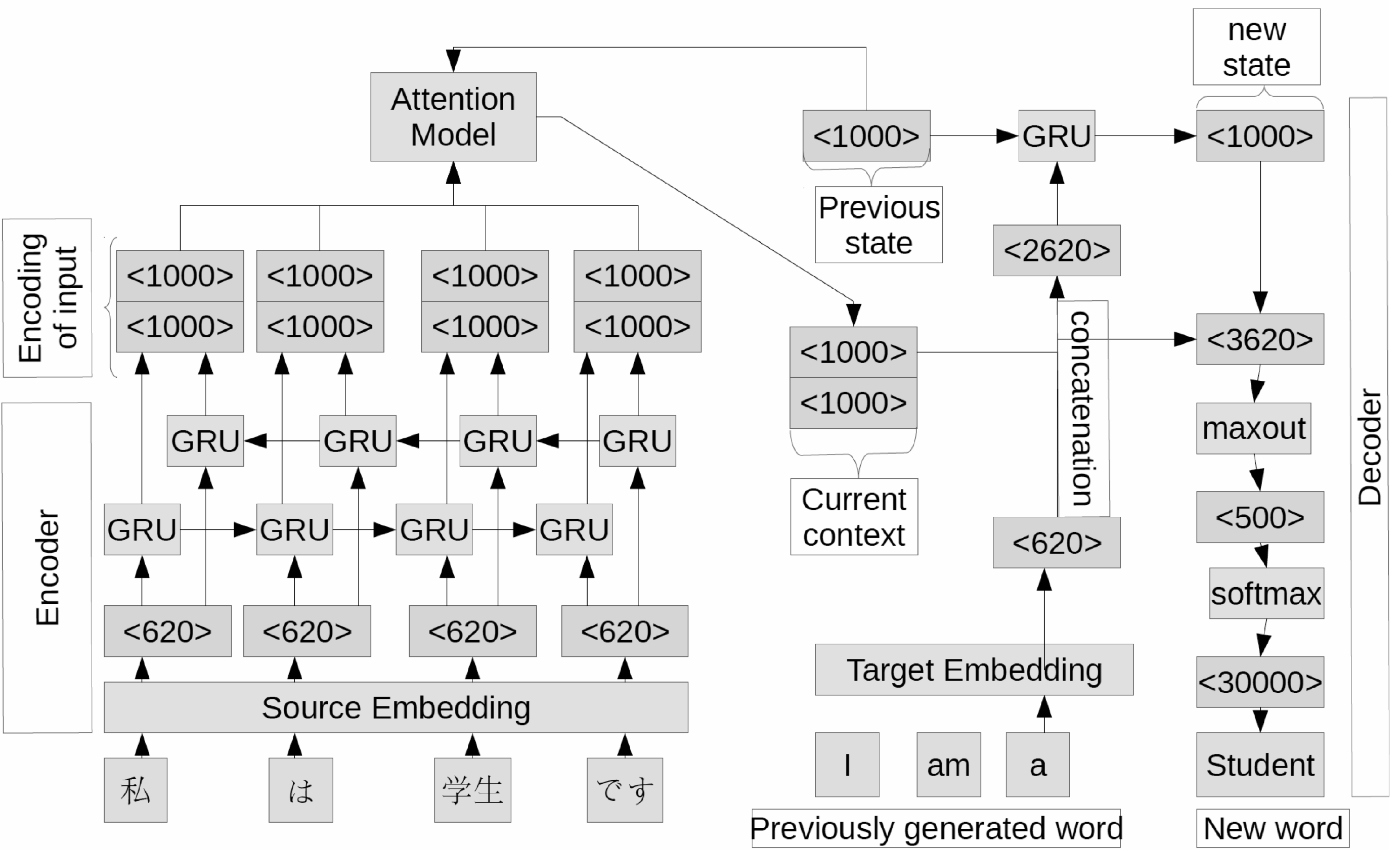}
     \end{center}
     \caption{\label{fig:rnnsearch} The architecture of the NMT system with attention,
as described in \cite{DBLP:journals/corr/BahdanauCB14}. The
notation ``\textless 1000\textgreater" means a vector of size 1000. The vector
sizes shown here are the ones suggested in the original paper.}
 \end{figure}
 
NMT is an end-to-end approach for translating from one language to another, which relies on deep learning to train a translation model \cite{DBLP:journals/corr/ChoMGBSB14,DBLP:journals/corr/SutskeverVL14,DBLP:journals/corr/BahdanauCB14}. The encoder-decoder model with attention \cite{DBLP:journals/corr/BahdanauCB14} is the most commonly used NMT architecture.
This model is also known as RNNsearch. Figure~\ref{fig:rnnsearch} describes the RNNsearch model \cite{DBLP:journals/corr/BahdanauCB14}, which takes in an input sentence $\mathbf{x} = \{x_1, ..., x_n\}$ and its translation $\mathbf{y} = \{y_1, ..., y_m\}$. The translation is generated as:
\begin{equation}
p(\mathbf{y}|\mathbf{x};\mathbf{\theta}) = \prod_{j=1}^{m} p(y_j|y_{<j}, \mathbf{x};\mathbf{\theta}),
\end{equation}
where $\mathbf{\theta}$ is a set of parameters, $m$ is the entire number of words in $\mathbf{y}$, $y_j$ is the current predicted word, and $y_{<j}$ are the previously predicted words.
Suppose we have a parallel corpus $C$ consisting of a set of parallel sentence pairs $(\mathbf{x, y})$. The training object is to  minimize the cross-entropy loss $L$ w.r.t $\mathbf{\theta}$:
\begin{equation}
L_\theta = \sum_{(\mathbf{x,y})\in\mathbf{C}}-\log p(\mathbf{y}|\mathbf{x}; \theta).
\end{equation}

The model consists of three main parts, namely, the encoder, decoder and attention model. 
The encoder uses an embedding mechanism to convert words into their continuous space representations. These embeddings by themselves do not contain information about relationships between words  and their positions in the sentence. Using a recurrent neural network (RNN) layer, gated recurrent unit (GRU) in this case, this can be accomplished. An RNN maintains a hidden state (also called a memory or history), which allows it to generate a continuous space representation for a word given all past words that have been seen. There are two GRU layers which encode forward and backward information. Each word $x_i$ is represented by concatenating the forward hidden state $\overrightarrow{h_i}$ and the backward one  $\overleftarrow{h_i}$ as $h_i = [\overrightarrow{h_i} ; \overleftarrow{h_i} ]$. In this way, the source sentence $\mathbf{x} = \{x_1, ..., x_n\}$ can be represented as $\mathbf{h} = \{h_1, ..., h_n\}$. By using both forward and backward recurrent information, one obtains a continuous space representation for a word given all words before as well as after it. 

The decoder is conceptually an RNN language model (RNNLM) with its own embedding mechanism, a GRU layer to remember previously generated words and a softmax layer to predict a target word. 
The encoder and decoder are coupled by using an attention mechanism, which computes a weighted average of the recurrent representations generated by the encoder thereby acting as a soft alignment mechanism.
This weighted averaged vector, also known as the context or attention vector, is fed to the decoder GRU along with the previously predicted word to produce a representation that is passed to the softmax layer to predict the next word. 
In equation, an RNN hidden state $s_j$ for time $j$ of the decoder is computed by:
\begin{equation}
s_j=f(s_{j-1}, y_{j-1}, c_j),
\end{equation}
where $f$ is an activation function of GRU, $s_{j-1}$ is the prvious RNN hidden state, $y_{j-1}$ is the previous word, $c_j$ is the context vector. $c_j$ is computed as a weighted sum of the encoder hidden states $\mathbf{h} = \{h_1, ..., h_n\}$, by using alignment weight $a_{ji}$:
\begin{eqnarray}
c_j=\sum_{j=1}^{n}{a_{ji}}{h_i}, \quad
a_{ji}=\frac{\mathrm{exp}(e_{ji})}{\sum_{k=1}^{m}\mathrm{exp}(e_{ki})}, \quad
e_{ji}=a(s_{j-1}, h_i),
\end{eqnarray}
where $a$ is an alignment model that scores the match level of the inputs around position $i$ and the output at position $j$.
The softmax layer contains a maxout layer which is a feedforward layer with max pooling. The maxout layer takes the recurrent hidden state generated by the decoder GRU, the previous word and the context vector to compute a final representation, which is fed to a simple softmax layer:
\begin{eqnarray}
P(y_j|y_{<j}, \mathbf{x}) = {\rm softmax} ({\rm maxout}(s_j, y_{j-1}, c_j)).
\end{eqnarray}

An abundance of parallel corpora are required to train an NMT system to avoid overfitting, due to the large amounts of parameters in the encoder, decoder, and attention model. This is the main bottleneck of NMT for low resource domains and languages.

\section{Domain Adaptation for SMT}
\label{sec:da-smt}

In SMT, many domain adaptation methods have been proposed to overcome the problem of the lack of substantial data in specific domains and languages. Most SMT domain adaptation methods can be broken down broadly into two main categories:

\subsection{Data Centric}
\label{sec:smt_data}
This category focuses on selecting or generating the domain-related data using existing in-domain data.

\romannumeral1) When there are sufficient parallel corpora from other domains, the main idea is to score the out-domain data using models trained from the in-domain and out-of-domain data and select training data from the out-of-domain data using a cut-off threshold on the resulting scores. LMs \cite{moore-lewis:2010:Short,axelrod-he-gao:2011:EMNLP,duh-EtAl:2013:Short},
as well as joint models \cite{hoang-simaan:2014:Coling,joty2015using}, and more recently convolutional neural network (CNN) models \cite{chen2016bilingual} can be used to score sentences.

\romannumeral2) When there are not enough parallel corpora, there are also studies that generate pseudo-parallel sentences using information retrieval \cite{utiyama-isahara:2003:ACL}, self-enhancing \cite{Lambert:2011:ITM:2132960.2132997} or parallel word embeddings \cite{marie-fujita:2017:Short}. Besides sentence generation, there are also studies that generate monolingual $n$-grams \cite{wang-EtAl:2014:EMNLP20142} and parallel phrase pairs \cite{chu:2015,wang-EtAl:2016:COLING5}.

Most of the data centric-based methods in SMT can be directly applied to NMT.  However, most of these methods adopt the criteria of data selection or generation that are not related to NMT. Therefore, these methods can only achieve modest improvements in NMT \cite{wang-EtAl:2017:Short3}.

\subsection{Model Centric} 

This category focuses on interpolating the models from different domains.

\romannumeral1) Model level interpolation. Several SMT models, such as LMs, translation models, and reordering models, individually corresponding to each corpus, are trained. These models are then combined to achieve the best performance \cite{Foster:2007:MAS:1626355.1626372,DBLP:conf/iwslt/BisazzaRF11,Niehues2012DetailedAO,sennrich-schwenk-aransa:2013:ACL2013,joty2015using,Imamura:DomainAdaptation2016-1}.

\romannumeral2) Instance level interpolation. Instance weighting has been applied to several natural language processing (NLP) domain adaptation tasks \cite{jiang-zhai:2007:ACLMain}, especially SMT \cite{matsoukas-rosti-zhang:2009:EMNLP,foster-goutte-kuhn:2010:EMNLP,Shah_amta2012,Mansour_2012,zhou2015domain}. They firstly score each instance/domain by using rules or statistical methods as a weight, and then train SMT models by giving each instance/domain the weight. An alternative way is to weight the corpora by data re-sampling \cite{shah2010translation,8211}.

For NMT, several methods have been proposed to interpolate model/data like SMT does.  For model-level interpolation, the most related NMT technique is model ensemble \cite{jean-EtAl:2015:ACL-IJCNLP}. For instance-level interpolation, the most related method is to assign a weight in NMT objective function \cite{costweighting,iwnmt}. However, the model structures of SMT and NMT  are quite different. SMT is a combination of several independent models; in comparison, NMT is an integral model itself. Therefore, most of these methods cannot be directly applied to NMT.

\section{Domain Adaptation for NMT}
\label{sec:da-nmt}

\subsection{Data Centric}
\subsubsection{Using Monolingual Corpora }
Unlike SMT, in-domain monolingual data cannot be used as an LM for conventional NMT directly, and many studies have been conducted for this.
G\"{u}l\c{c}ehre et al. \shortcite{DBLP:journals/corr/GulcehreFXCBLBS15} train an RNNLM on monolingual data, and fuse the RNNLM and NMT models. Currey et al.
\shortcite{currey-micelibarone-heafield:2017:WMT} copy the target monolingual data to the source side and use the copied data for training NMT. 
Domhan and Hieber \shortcite{domhan-hieber:2017:EMNLP2017} propose using target monolingual data for the decoder with LM and NMT multitask learning.  Zhang and Zong \shortcite{zhang-zong:2016:EMNLP2016} use source side monolingual data to strengthen the NMT encoder via multitask learning for predicting both translation and reordered source sentences.  Cheng et al. \shortcite{cheng-EtAl:2016:P16-1} use both source and target monolingual data for NMT through reconstructing the monolingual data by using NMT as an autoencoder.


\subsubsection{Synthetic Parallel Corpora Generation}
\begin{figure}[t]
    \centering
  \centerline{\includegraphics[width=0.6\hsize]{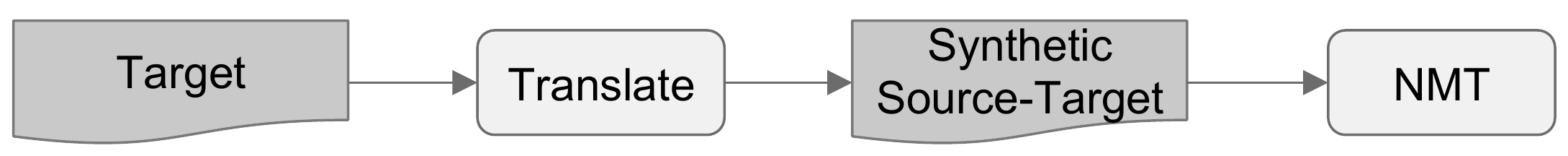}}
        \caption{Synthetic data generation for NMT \cite{sennrich-haddow-birch:2016:P16-11}.}
      \label{fig:synthetic}
\end{figure}

As NMT itself has the ability of learning LMs, target monolingual data also can be used for the NMT system to strengthen the decoder after back translating target sentences to generate a synthetic parallel corpus \cite{sennrich-haddow-birch:2016:P16-11}.
Figure \ref{fig:synthetic} shows the flowchart of this method. 
It has also been shown that synthetic data generation is very effective for domain adaptation using either the target side monolingual data \cite{sennrich-haddow-birch:2016:P16-12},  the source side monolingual data \cite{zhang-zong:2016:EMNLP2016}, or both \cite{DBLP:journals/corr/ParkSY17}.

\subsubsection{Using Out-of-Domain Parallel Corpora}
With both in-domain and out-of-domain parallel corpora, it is ideal to train a mixed domain MT system that can improve in-domain translation while do not decrease the quality of out-of-domain translation. We categorize these efforts as {\it multi-domain} methods, which have been successfully developed for NMT. In addition, the idea of data selection from SMT also have been developed for NMT.

{\bf Multi-Domain}
The {\it multi-domain} method in Chu et al. \shortcite{chu:2017:ACL} is originally motivated by Sennrich et al. \shortcite{sennrich-haddow-birch:2016:N16-1}, which 
uses tags to control the politeness of NMT. The overview of this method is shown in the dotted section in Figure \ref{mlnmt}. 
In this method, the corpora of multiple domains are concatenated  with two small modifications: 
\begin{itemize}
\item Appending the domain tag ``\textless2domain\textgreater" to the source sentences of the respective corpora.
This primes the NMT decoder to generate sentences for the specific domain. 
\item Oversampling the smaller corpus so that the training procedure pays equal attention to each domain.
\end{itemize}

Sajjad et al. \shortcite{DBLP:journals/corr/abs-1708-08712} further compare different methods for training a multi-domain system. In particular, they compare {\it concatenation} that simply concatenates the multi-domain corpora, {\it staking} that iteratively trains the NMT system on each domain corpus, {\it selection} that selects a set of out-of-domain data which is close to the in-domain data, and {\it ensemble} that ensembles the multiple NMT models trained independently. They find that fine tuning the concatenation system on in-domain data shows the best performance. Britz  et al. \shortcite{britz-le-pryzant:2017:WMT} compare the {\it multi-domain} method with a discriminative method (see Section \ref{sec:arch} for details). They show that the discriminative method performs better than the {\it multi-domain} method.

\begin{figure}[t]
    \centering
  \centerline{\includegraphics[width=1\hsize]{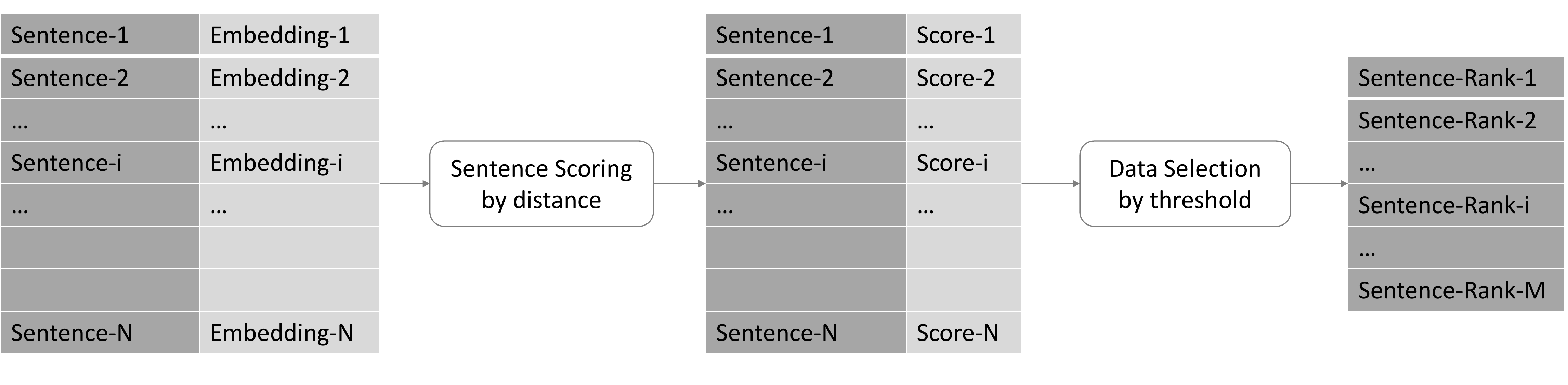}}
        \caption{Data selection for NMT \cite{wang-EtAl:2017:Short3}.}
      \label{fig:selection}
\end{figure}

{\bf Data Selection} 
As mentioned in the SMT section (Section \ref{sec:smt_data}), the data selection methods in SMT can improve NMT performance modestly, because their criteria of data selection are not very related to NMT \cite{wang-EtAl:2017:Short3}. To address this problem, Wang et al. \shortcite{wang-EtAl:2017:Short3} exploit the internal embedding of the source sentence in NMT, and use the sentence embedding similarity to select the sentences that are close to in-domain data from out-of-domain data (Figure \ref{fig:selection}). Van der Wees et al. \shortcite{vanderwees-bisazza-monz:2017:EMNLP2017} propose a dynamic data selection method, in which they change the selected subset of training data among different training epochs for NMT. They show that gradually decreasing the training data based on the in-domain similarity gives the best performance.

Although all the data centric methods for NMT are complementary to each other in principle, there are no studies that try to combine these methods, which is considered to be one future direction.

\subsection{Model Centric} 
\subsubsection{Training Objective Centric} 
The methods in this section change the training functions or procedures for obtaining an optimal in-domain training objective.

\begin{figure}[t]
    \centering
  \centerline{\includegraphics[width=1\hsize]{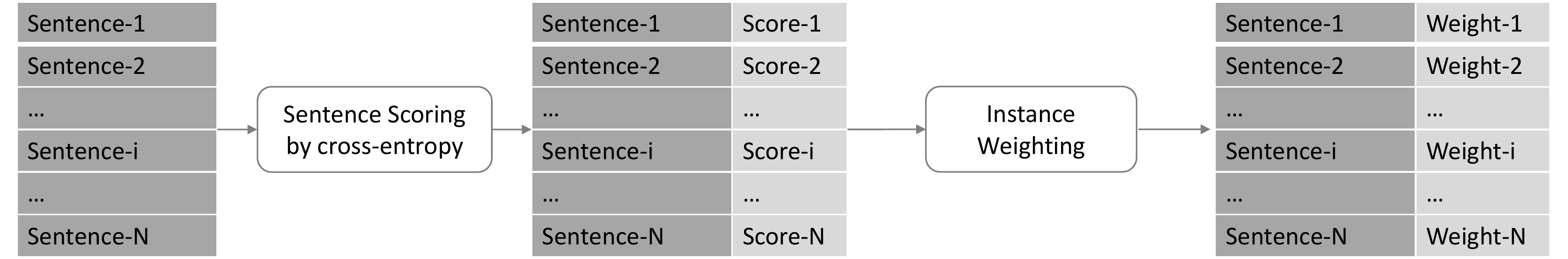}}
        \caption{Instance weighting for NMT \cite{iwnmt}. }
      \label{fig:instance}
\end{figure}

{\bf Instance/Cost Weighting}
The main challenge for instance weighting in NMT is that NMT is not a linear model or a combination of linear models, which means the instance weight cannot be integrated into NMT directly. 
There is only one work concerning instance weighting in NMT \cite{iwnmt}. They set a weight for the objective function, and this weight is learned from the cross-entropy by an in-domain LM and an out-of-domain LM \cite{axelrod-he-gao:2011:EMNLP} (Figure \ref{fig:instance}). Instead of instance weighting,  Chen et al. \shortcite{costweighting} modify the NMT cost function with a domain classifier. The output probability of the domain classifier is transferred into the domain weight. This classifier is trained using development data. Recently, Wang et al. \shortcite{8360031} proposed a joint framework of sentence selection and weighting for NMT.

\begin{figure}[t]
    \centering
  \centerline{\includegraphics[width=0.8\hsize]{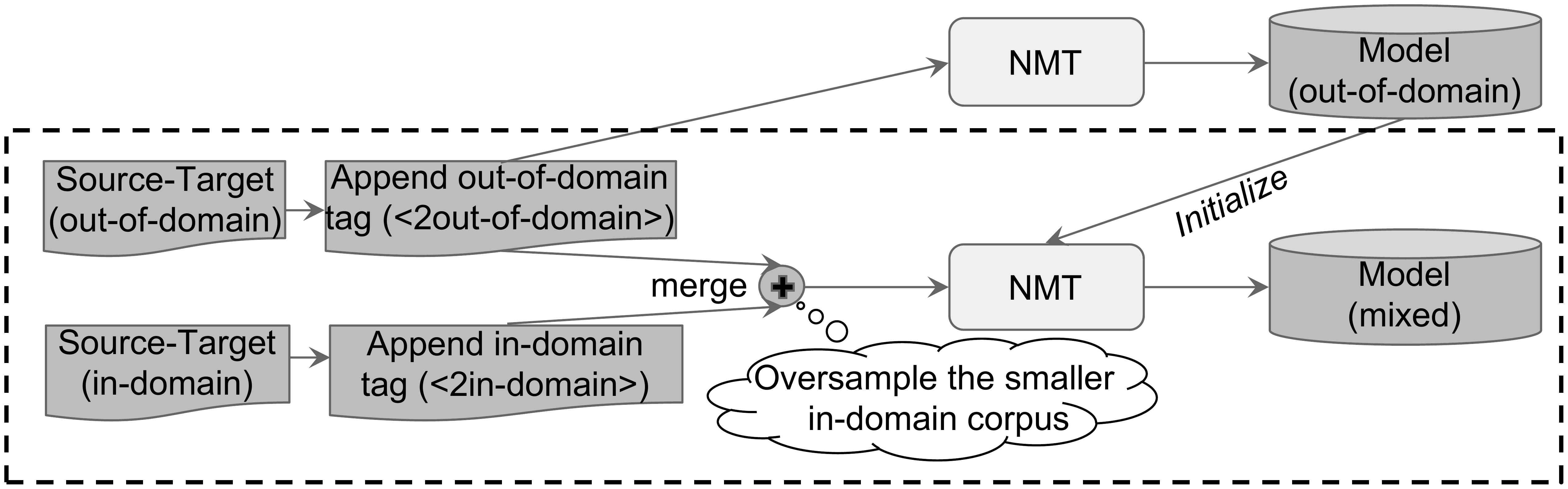}}
      \caption{Mixed fine tuning with domain tags for domain adaptation \cite{chu:2017:ACL}. The section in the dotted rectangle denotes the {\it multi-domain} method . }
        \label{mlnmt}
\end{figure}

{\bf Fine Tuning }
{\it Fine tuning} is the conventional way for domain adaptation \cite{luong2015stanford,sennrich-haddow-birch:2016:P16-11,domspec,domfast}. 
In this method,  an NMT system on a resource 
rich out-of-domain corpus is trained until convergence, and then its parameters are fine tuned  on a resource poor in-domain corpus. 
Conventionally, fine tuning is applied on in-domain parallel corpora. Varga et al. \shortcite{varga:2017} apply it on parallel sentences extracted from comparable corpora. Comparable corpora have been widely used for SMT by extracting parallel data from them \cite{chu:2015}. To prevent degradation of out-of-domain translation after fine tuning on in-domain data, Dakwale and Monz \shortcite{dakw:17fine} propose an extension of fine tuning that keeps the distribution of the out-of-domain model  based on knowledge distillation \cite{Hinton:nips2015}.

{\bf Mixed Fine Tuning }
This method is a combination of the {\it multi-domain} and {\it fine tuning} methods (Figure \ref{mlnmt}).
The training procedure is as follows:
\begin{enumerate}
\item Train an NMT model on out-of-domain data until convergence.
\item Resume training the NMT model from step 1 on a mix of in-domain and out-of-domain data (by oversampling the in-domain data) until convergence.
\end{enumerate}
Mixed fine tuning addresses the overfitting problem of fine tuning due to the small size of the in-domain data.
It is easier to train a good model with out-of-domain data, compared to training a multi-domain model. Once we obtained good model parameters, we can use these parameters for fine tuning on the mixed domain data to obtain better performance for the in-domain model. In addition, mixed fine tuning is faster than multi-domain because training an out-of-domain model convergences faster than training a multi-domain model, which also convergences very fast in fine tuning on the mixed domain data.
Chu et al. \shortcite{chu:2017:ACL} show that mixed fine tuning works better than both {\it multi-domain} and {\it fine tuning}. 
In addition, mixed fine tuning has the similar effect as the ensembling method in Dakw and Monz \shortcite{dakw:17fine}, which does not decrease the out-of-domain translation performance.

 
{\bf Regularization} Barone et al. \shortcite{micelibarone-EtAl:2017:EMNLP2017} also realize the overfitting problem during fine tuning. Their strategy to address this problem is to explore regularization techniques such as dropout and L2-regularization. In addition, they also propose {\it tuneout} that is a variant of dropout for regularization. We think that mixed fine tuning and regularization techniques are complementary to each other.
 
\subsubsection{Architecture Centric }
\label{sec:arch}
The methods in this section change the NMT architecture for domain adaptation. 

{\bf Deep Fusion}
One technique of adaptation with in-domain monolingual data is to train an in-domain RNNLM for the NMT decoder and combine it (also known as fusion) with an NMT model \cite{DBLP:journals/corr/GulcehreFXCBLBS15}. Fusion can either be shallow or deep. Formally, deep fusion indicates that the LM and NMT are integrated as a single decoder (i.e., integrating the RNNLM into the NMT architecture). Shallow fusion indicates that the scores of the LM and NMT are considered together (i.e., rescoring the NMT model with the RNNLM model).

In deep fusion,  the RNNLM and the decoder of the NMT are integrated by concatenating their hidden states. When computing the output probability of the next word, the model is fine tuned to use the hidden states of both the RNNLM and NMT models. Domhan and Hieber \shortcite{domhan-hieber:2017:EMNLP2017} propose a method similar to the deep fusion method \cite{DBLP:journals/corr/GulcehreFXCBLBS15}. However, unlike training the RNNLM and NMT model separately \cite{DBLP:journals/corr/GulcehreFXCBLBS15}, Domhan and Hieber \shortcite{domhan-hieber:2017:EMNLP2017} train RNNLM and NMT models  jointly.

\begin{figure}[t]
    \centering
  \centerline{\includegraphics[width=0.6\hsize]{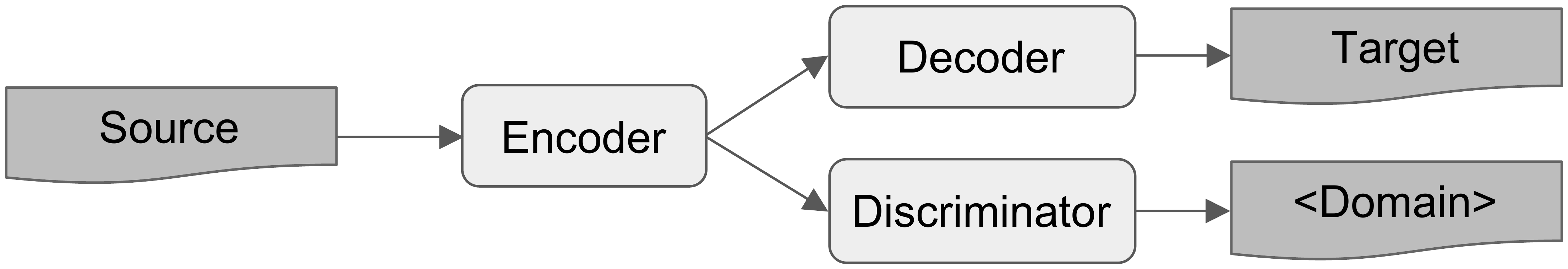}}
        \caption{Domain discriminator \cite{britz-le-pryzant:2017:WMT}.}
      \label{fig:discriminator}
\end{figure}

{\bf Domain Discriminator}
To leverage the diversity of information in multi-domain corpora, Britz et al. \shortcite{britz-le-pryzant:2017:WMT} propose a discriminative method. In their discriminative method, they add a feed-forward network (FFNN) as a discriminator on top of the encoder that uses the attention to predict the domain of the source sentence. The discriminator is optimized jointly with
the NMT network. 
Figure \ref{fig:discriminator} shows an overview of this method.


{\bf Domain Control }
Besides using domain tokens to control the domains,  Kobus et al. \shortcite{domcont} propose to append word-level features to the embedding layer of NMT to control the domains. In particular, they append a domain tag to each word. They also propose a term frequency - inverse document frequency (tf-idf) based method to predict the domain tag for input sentences.

\subsubsection{Decoding Centric}
\begin{figure}[t]
    \centering
  \centerline{\includegraphics[width=0.6\hsize]{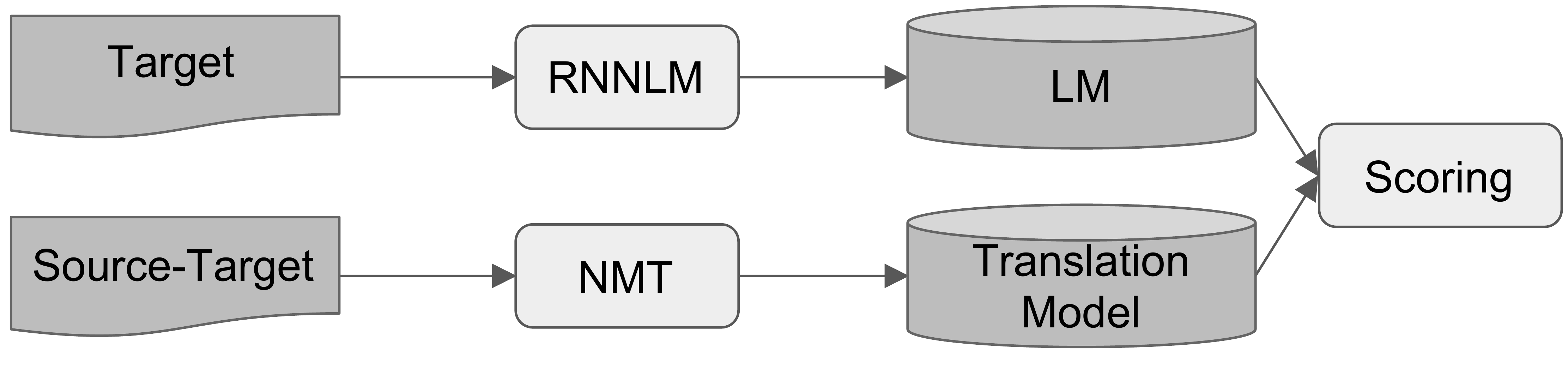}}
        \caption{LM shallow fusion \cite{DBLP:journals/corr/GulcehreFXCBLBS15}.}
      \label{fig:lm_sf}
\end{figure}

Decoding centric methods focus on the decoding algorithm for domain adaptation, which are essentially complementary to the other model centric methods.

{\bf Shallow Fusion }
Shallow fusion is an approach where LMs are trained on large monolingual corpora, following which they are combined with a previously trained NMT model \cite{DBLP:journals/corr/GulcehreFXCBLBS15}. 
In the shallow fusion \cite{DBLP:journals/corr/GulcehreFXCBLBS15}, the next word hypotheses generated by an NMT model is rescored by the weighted sum of the NMT and RNNLM probabilities (Figure \ref{fig:lm_sf}).

{\bf Ensembling}
Freitag and Al-Onaizan \shortcite{domfast} propose to ensemble the out-of-domain domain and the fine tuned in-domain models. Their motivation is exactly the same as the work of Dakwale and Monz \shortcite{dakw:17fine}, which is preventing degradation of out-of-domain translation after fine tuning on in-domain data. 

{\bf Neural Lattice Search } Khayrallah et al. \shortcite{khayrallah-EtAl:2017:I17-2} propose a stack-based decoding algorithm over word lattices, while the lattices are generated by SMT \cite{dyer-muresan-resnik:2008:ACLMain}. In their domain adaptation experiments, they show that stack-based decoding is better than conventional decoding.





\section{Domain Adaptation in Real-World Scenarios}
\label{sec:real}

\begin{figure}[t]
    \centering
  \centerline{\includegraphics[width=0.6\hsize]{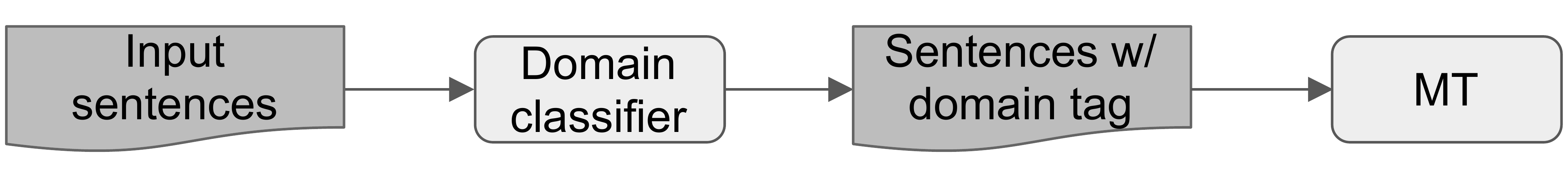}}
        \caption{Domain adaptation in an input domain unknown scenario.}
      \label{fig:real}
\end{figure}

A domain adaptation method should be adopted according to the certain scenarios. For example, when there are some pseudo parallel in-domain data in the out-of-domain data, sentence selection is preferred; when only additional monolingual data is available, LM and NMT fusion can be adopted. In many cases, both out-of-domain parallel data and monolingual in-domain data are available, making the combination of different methods possible. Chu et al. \shortcite{Chu:2018:JIP} conduct a study that applys mixed fine tuning \cite{chu:2017:ACL} on synthetic parallel data \cite{sennrich-haddow-birch:2016:P16-11}, which shows better performance than either method. Therefore, we do not recommend any particular techniques in this paper but recommend readers to choose the best method for their own scenarios.

Most of the above domain adaptation studies assume that the domain of the data is given. However, in a practical view such as an online  translation engine, the domain of the sentences input by the users are not given. For such scenario, predicting the domains of the input sentences is crucial for good translation. 
To address this problem,  a common method in SMT is to firstly classify the domains and then translate input sentences in classified domains using corresponding models \cite{huck2015mixed}. Xu et al. \shortcite{xu2007domain} perform domain classification for a Chinese-English translation task. The classifiers operate on whole documents rather than on individual sentences, using LM interpolation and vocabulary similarities. Huck et al. \shortcite{huck2015mixed} extend the work of Xu et al. \shortcite{xu2007domain} on the sentence level. They use LMs and maximum entropy classifiers to predict the target domain. Banerjee et al. \shortcite{banerjee2010combining} build a support vector machine classifier using tf-idf features over bigrams of stemmed content words. Classification is carried out on the level of individual sentences. Wang et al. \shortcite{wang2012improved} rely on averaged perceptron classifiers with various phrase-based features. 

For NMT, Kobus et al. \shortcite{domcont} propose an NMT domain control method, by appending either domain tags or features to the word embedding layer of NMT. They adopt an in-house classifier to distinguish the domain information. Li et al. \shortcite{DBLP:journals/corr/LiZZ16d} propose to search similar sentences in the training data using the test sentence as a query, and then fine tune the NMT model using the retrieved training sentences for translating the test sentence. Farajian et al. \shortcite{farajian-EtAl:2017:WMT} follow the strategy of Li et al. \shortcite{DBLP:journals/corr/LiZZ16d}, but propose to dynamically set the hyperparameters (i.e., learning rate and number
of epochs) of the learning algorithm based on the similarity of the input sentence and the retrieved sentences for updating the NMT model. Figure \ref{fig:real} shows an overview of domain adaptation for MT in the input domain unknown scenario.

\section{Future Directions}
\label{sec:future}
\subsection{Domain Adaptation for State-of-the-art NMT Architectures} 
Since the success of RNN based NMT \cite{DBLP:journals/corr/ChoMGBSB14,DBLP:journals/corr/SutskeverVL14,DBLP:journals/corr/BahdanauCB14}, 
other architectures of NMT have been developed. One representative architecture is CNN based NMT \cite{DBLP:journals/corr/GehringAGYD17}. Compared to RNN based models, CNN based models can be computed fully parallel during training and are much easier to optimize.
Another representative architecture is the {\it Transformer}, which is based on attention only \cite{NIPS2017_7181}. It has been shown that CNN based NMT and the {\it Transformer} significantly outperform the state-of-the-art  RNN based NMT model of Wu et al. \shortcite{DBLP:journals/corr/WuSCLNMKCGMKSJL16} in both the translation quality and speed perspectives. However, currently, most of the domain adaptation studies for NMT are based on the RNN based model \cite{DBLP:journals/corr/BahdanauCB14}.
The research of domain adaptation techniques for these latest state-of-the-art NMT models is obviously an important future direction.

\subsection{Domain Specific Dictionary Incorporation} 
How to use external knowledge such as dictionaries and knowledge bases for NMT remains a big research question.
In domain adaptation, the use of domain specific dictionaries is a very crucial problem. In the practical perspective, many translation companies have created domain specific dictionaries but not domain specific corpora. If we can study a good way to use domain specific dictionaries, it will significantly promote the practical use of MT. There are some studies that try to use dictionaries for NMT, but the usage is limited to help low frequent or rare word translation \cite{arthur-neubig-nakamura:2016:EMNLP2016,DBLP:journals/corr/ZhangZ16c}. 
Arcan and Buitelaar \shortcite{DBLP:journals/corr/abs-1709-02184} use  a domain specific dictionary for  terminology translation, but they simply apply the unknown word replacement method proposed by Luong et al. \shortcite{luong-EtAl:2015:ACL-IJCNLP}, which suffers from noisy attention.

\subsection{Multilingual and Multi-Domain Adaptation}
It may not always be possible to use an out-of-domain parallel corpus in the same language pair and thus it is important to use data from other languages \cite{gnmt16multi}.
This approach is known as cross-lingual transfer learning, which transfers NMT model parameters among multiple languages. It is known that a multilingual model, which relies on parameter sharing, helps in improving the translation quality for low resource languages 
especially when the target language is the same \cite{DBLP:conf/emnlp/ZophYMK16}.
There are studies where either multilingual \cite{DBLP:journals/corr/FiratCB16,TACL1081} or multi-domain models \cite{DBLP:journals/corr/abs-1708-08712} are trained, but none that attempt to package multiple language pairs and multiple domains into a single translation system. Even if out-of-domain data in the same language pair exists, it is possible that using both multilingual and multi-domain data can boost the translation performance. Therefore, we think that multilingual and multi-domain adaptation for NMT can be another future direction. Chu and Dabre \shortcite{chu:2018:NLP2018} conduct a preliminary study for this topic.

\subsection{Adversarial Domain Adaptation and Domain Generation}
Generative adversarial networks are a class of artificial intelligence algorithms used in unsupervised machine learning,  which are introduced by \cite{goodfellow2014generative}. Adversarial methods have become popular in domain adaptation \cite{JMLR:v17:15-239}, which minimize an approximate domain discrepancy distance through an adversarial objective with respect to a domain discriminator \cite{DBLP:journals/corr/TzengHSD17}. They have been applied to domain adaptation tasks in computer vision and machine learning \cite{DBLP:journals/corr/TzengHSD17,DBLP:journals/corr/abs-1711-02536,DBLP:journals/corr/abs-1711-08561,DBLP:journals/corr/ZhaoZWCMG17,pei2018multi}. Recently, some of the adversarial methods began to be introduced into some NLP tasks \cite{liu-qiu-huang:2017:Long,chen-EtAl:2017:Long2} and NMT \cite{britz-le-pryzant:2017:WMT}. 

Most of the existing methods focus on adapting from a general domain into a specific domain. In the real scenario, training data and test data have different distributions and the target domains are sometimes unseen. Irvine et al. \shortcite{Q13-1035} analyze the translation errors in such scenarios. Domain generalization aims to apply knowledge gained from labeled source domains to unseen target domains \cite{dg}. It provides a way to match the distribution of training data and test data  in real-world MT, which may be a future trend of  domain adaptation for NMT. 

\section{Conclusion}
\label{sec:conl}
Domain adaptation for NMT is a rather new but very important research topic to promote MT for practical use. In this paper, we gave the first comprehensive review of the techniques mainly being developed in the last two years. We compared domain adaptation techniques for NMT with the techniques being studied in SMT, which has been the main research area in the last two decades. In addition, we outlooked the future research directions. Connecting  domain adaptation techniques in NMT to the techniques in general NLP, computer vision and machine learning is our future work. We hope that this survey paper could significantly promote the research in domain adaptation for NMT.


\section*{Acknowledgement}
\noindent This work was supported by Grant-in-Aid for Research Activity Start-up \#17H06822, JSPS. We are very appreciated to Dr. Raj Dabre for the deep discussion of the structure for this paper. We also thank
the anonymous reviewers for their insightful comments.


\bibliographystyle{acl}
\bibliography{coling2018}

\end{document}